\documentclass{article}
\usepackage{spconf,amsmath,graphicx}
\usepackage{enumitem}
\setlist{nosep, leftmargin=14pt}
\usepackage{xcolor}

\usepackage{adjustbox}
\usepackage{parskip}
\usepackage{multirow}
\usepackage{booktabs}
\usepackage{amssymb} 
\usepackage{hyperref}

\title{Rethinking Intermediate Layers design in Knowledge Distillation for Kidney and  Liver Tumor Segmentation}
\name{Vandan Gorade$^{1}$, Sparsh Mittal$^2$, Debesh Jha$^1$, Ulas Bagci$^1$
\address{$^1$ Machine \& Hybrid Intelligence Lab, Department of Radiology, Northwestern University,  USA \\
$^2$ Mehta Family School of DS\&AI, Indian Institute of Technology, Roorkee, India\\
vandan.gorade@northwestern.edu, sparsh.mittal@ece.iitr.ac.in, \{debesh.jha, ulas.bagci\}@northwestern.edu}
\thanks{Sparsh and Ulas are corresponding authors. The project is supported by NIH funding: R01-CA246704,  R01-CA240639, U01 DK127384-02S1, and U01-CA268808. The computing system used for this research was supported by IIT Roorkee under the grant FIG-100874.}}

\begin{document}

\maketitle

\textbf{Abstract.} Knowledge distillation (KD) has demonstrated remarkable success across various domains, but its application to medical imaging tasks, such as kidney and liver tumor segmentation, has encountered challenges. Many existing KD methods are not specifically tailored for these tasks. Moreover, prevalent KD methods often lack a careful consideration of `what' and `from where' to distill knowledge from the teacher to the student. This oversight may lead to issues like the accumulation of training bias within shallower student layers, potentially compromising the effectiveness of KD. To address these challenges, we propose Hierarchical Layer-selective Feedback Distillation (HLFD). HLFD strategically distills knowledge from a combination of middle layers to earlier layers and transfers final layer knowledge to intermediate layers at both the feature and pixel levels. This design allows the model to learn higher-quality representations from earlier layers, resulting in a robust and compact student model. Extensive quantitative evaluations reveal that HLFD outperforms existing methods by a significant margin. For example, in the kidney segmentation task, HLFD surpasses the student model (without KD) by over 10\%, significantly improving its focus on tumor-specific features. From a qualitative standpoint, the student model trained using HLFD excels at suppressing irrelevant information and can focus sharply on tumor-specific details, which opens a new pathway for more efficient and accurate diagnostic tools. Code is available \href{https://github.com/vangorade/RethinkingKD_ISBI24}{here}.


\section{Introduction}
Tumor segmentation in medical imaging enables clinicians to accurately identify, assess, and manage malignancies. Leveraging neural networks, we achieve automated, high-fidelity delineation of tumor boundaries in various imaging modalities, including CT scans and MRIs \cite{abdelrahman2022kidney,heller2019kits19,gorade2024synergynet}. This technological breakthrough elevates diagnostic accuracy and efficiency and streamlines treatment planning \cite{meyer2018survey}, ultimately leading to enhanced patient care and outcomes. 

Significant challenges persist despite the remarkable successes of deep-learning models in medical image segmentation~\cite{gorade2024harmonized,jha2024ct, Pal_2023_ICCV,10229358}. These models demand extensive datasets and substantial computational resources, making deployment on resource-limited devices a hurdle. Furthermore, the diversity in tumor appearances, irregular sizes, unpredictable locations, and variations amplifies segmentation complexity. To address these challenges, researchers are exploring innovative strategies. For instance, lightweight networks~\cite{ge2020survey,emara2019liteseg, jha2021nanonet, jha2021real} have been explored for real-time semantic segmentation, and recent works have delved into real-time medical image segmentation. However, model simplification may hurt predictive performance. Knowledge distillation (KD) \cite{gou2021knowledge} has emerged as a valuable approach, facilitating knowledge transfer from the larger `teacher' models to the leaner `student' models. 

Existing works \cite{hinton2015distilling,park2019relational,yang2020knowledge,tian2019contrastive,liu2019structured,ji2021show,gorade2023pacl} aim to enhance the final representations of the student model by minimizing the difference in softmax representations between the teacher and student models. However, this supervisory signal originates solely from the final student layer. Hence, it tends to attenuate with each layer during backpropagation, accumulating training bias within the shallower student layers. This impairs the efficacy of knowledge transfer. Other works~\cite{romero2014fitnets,zagoruyko2016paying,passalis2018learning,qin2021efficient} focus on improving the alignment of latent feature maps by mimicking intermediate representations. These intermediate representations serve as solid indicators that facilitate learning the final representation. However, when we replicate intermediate representations, we are limited to capturing the knowledge acquired by that specific layer, potentially missing out on global information. Recognizing this limitation,  capturing features from terminal representation at an earlier stage emerges as a valuable strategy~\cite{ijcai2023p108,luo2023knowledge}. However, these methods give suboptimal results where boundary, shape, texture information, and a combination of low-level features are essential, not only high-level class information. 
\begin{figure*}[!t]
\centering
\def\svgwidth{\columnwidth}
\includegraphics[width=0.82\textwidth]{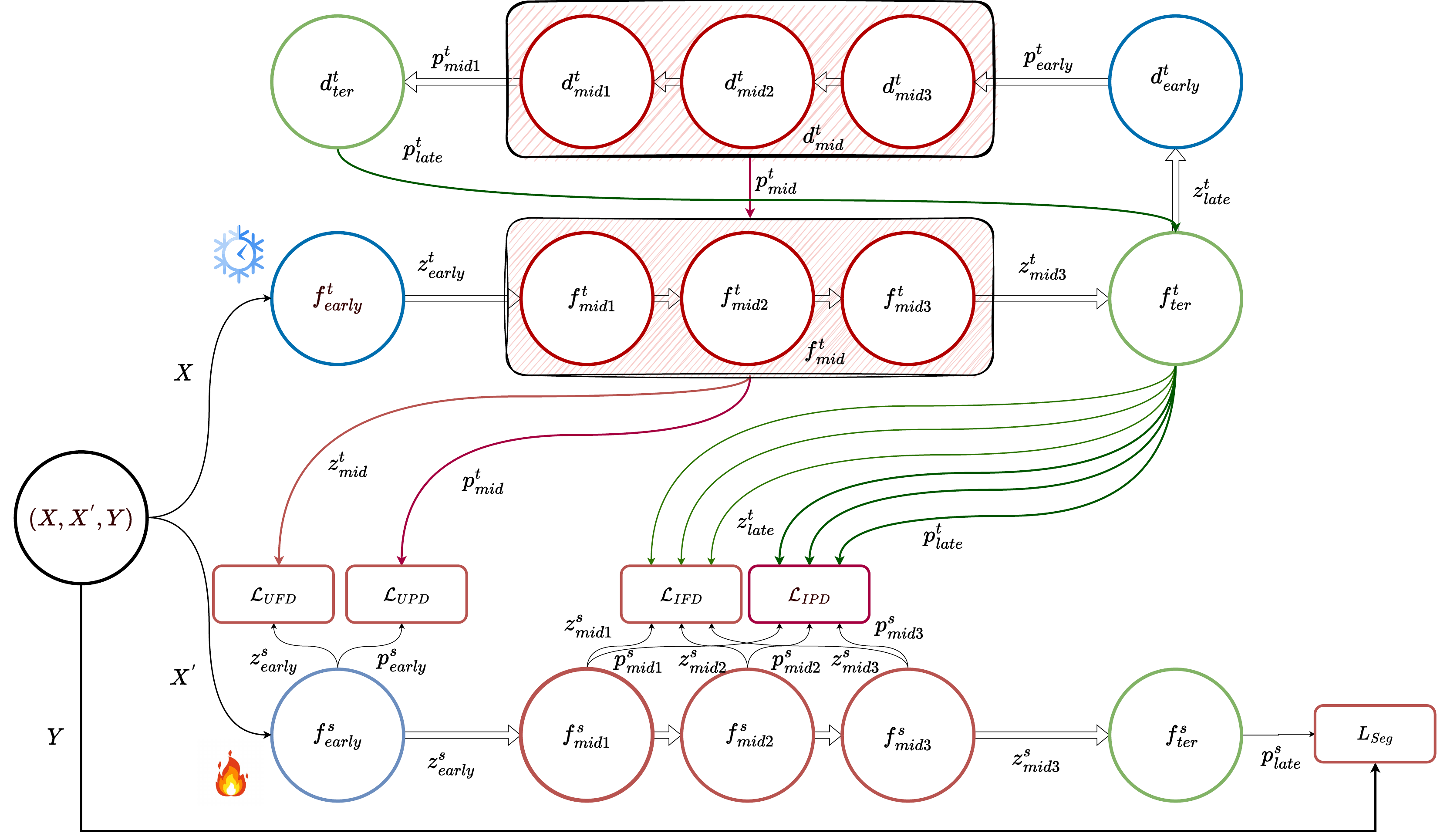}
\caption{The input X and augmentation $X'$, undergo encoding by both a pre-trained teacher encoder and a randomly initialized student encoder, resulting in representations $z^{t}_{early}$, $z^{t}_{\text{mid}j}$, $z^{t}_{ter}$, and $z^{s}_{early}$, $z^{s}_{\text{mid}j}$, $z^{s}_{ter}$, respectively. These representations contribute to feature-level loss functions, $\mathcal{L}_{UFD}$ and $\mathcal{L}_{IFD}$. Additionally, the teacher decoder decodes $z^{t}_{ter}$, producing representations $p^t{early}$, $p^t_{\text{mid}j}$, $p^t_{ter}$, which are utilized in pixel-level loss functions, $\mathcal{L}_{UPD}$ and $\mathcal{L}_{IPD}$. The training process is further enhanced with the inclusion of a supervised focal dice loss ($\mathcal{L}_{seg}$).
}
\label{fig: workflow_1}
\end{figure*}
We have introduced hierarchical layer-selective feedback distillation (HLFD) to address these challenges. HLFD comprises Feature-level LFD (FLFD) and Pixel-level LFD (PLFD). FLFD, in turn, includes Unified Feature-level Distillation (UFD) for unified representations and Individual Feature-level Distillation (IFD) for middle-to-early and later-to-middle layer distillation. PLFD, on the other hand, involves Unified Pixel-level Distillation (UPD) and Individual Pixel-level  Distillation (IPD), which transfers pixel-level knowledge from the teacher decoder to the student through interpolated features. HLFD integrates both FLFD and PLFD in a multi-task fashion, promoting simultaneous learning of feature-level and pixel-level representations. Our contributions are as follows, 
\begin{itemize}
    \item We rethink the design of layers in the context of distillation and introduce the Hierarchical Layer-selective Feedback Distillation (HLFD) framework. 
    \vspace{-0.2pt}
    \item We demonstrate HLFD's capability to capture tumor-specific details from early layers while effectively suppressing irrelevant information flow.
    \vspace{-0.2pt}
    \item Extensive experiments conducted on kidney and liver tumor segmentation tasks establish that our proposed method attains state-of-the-art (SOTA) results
    
\end{itemize}


\section{Proposed Method}
\subsection{Feature-level Layer-selective Feedback Distillation (FLFD)}
Given an input $X$, transformations occur through both the pre-trained teacher encoder $f^{t}_i$ and the random student encoder $f^{s}_i$, denoted by \textit{i} for the number of blocks. This yields early representations $z^{t}_{early}$ and $z^{s}_{early}$, intermediate representations $z^{t}_{\text{mid}j}$ and $z^{s}_{\text{mid}j}$ (where \textit{j} is the number of middle layers), and terminal representations $z^{t}_{late}$ and $z^{s}_{late}$. These representations form the foundation of our framework. We propose an  FLFD loss,  defined as, $\mathcal{L}_{F} = \mathcal{L}_{UFD} +  \mathcal{L}_{IFD}$. These components are defined below.

\textbf{Unified Feature-level Distillation (UFD).} 
Within this framework, we introduce the concept of distilling the attentive
knowledge from the teacher's unified representation of middle layers, $z^t_{mid}$, to
the student’s early representation, $z^s_{early}$. To achieve this, we propose the following loss function. 
\begin{align}
    \mathcal{L}_{UFD} &= \frac{\| \mathcal{A}(z^{s}_{early}) \|}{\|\mathcal{A}(z^{s}_{early}) \|_{2}} - \frac{\| \mathcal{A}(z^{t}_{mid}) \|}{\|\mathcal{A}(z^{t}_{mid})\|_{2}}
    \label{eq:ufd}
\end{align}
To achieve $z^t_{mid}$, we perform interpolation on the middle layers with the larger feature maps to ensure their spatial dimensions match the smallest among them. Next, we concatenate all these interpolated representations along the channel dimension. Finally, operation A(.) is employed first to rescale the student's representation $z^{s}_{early}$ to match the spatial dimension of
the teacher's $z^{t}_{mid}$. Additionally, channel normalization is applied to the rescaled student representation, assuming that the absolute value of a neuron activation signifies its importance.

\textbf{Individual Feature-level Distillation (IFD).} Within this framework, we introduce the concept of distilling the attentive knowledge from the teacher's late representation, $z^t_{late}$, to each student's middle layers or intermediate representation, $z^{s}_{\text{mid}_j}$. To achieve this, we propose the following loss function. 
\begin{align}
    \mathcal{L}_{IFD} = \sum^{N}_{j=1} \frac{\| \mathcal{A}(z^{s}_{\text{mid}_j}) \|}{\|\mathcal{A}(z^{s}_{\text{mid}_j}) \|_{2}} - \frac{\| \mathcal{A}(z^{t}_{late}) \|}{\|\mathcal{A}(z^{t}_{late})\|_{2}}
    \label{eq:ifd}
\end{align}
Here, the operation $\mathcal{A(.)}$ is same as in Eq.\ref{eq:ufd}.

\subsection{Pixel-level Layer-selective Feedback Distillation (PLFD):} 
In contrast to feature-level distillation, pixel-level segmentation-map distillation is geared toward conveying pixel-wise predictions. In practice, we distill pixel-level maps generated by the teacher's decoder to interpolated student maps. First, the teacher encoder output $z^t_{late}$ is passed through pre-trained teacher decoder $d^t_i$ resulting in early predictive map $p^t_{early}$, intermediate predictive map $p^t_{\text{mid}_j}$ and terminal predictive map $p^t_{late}$. For students, we used an interpolated representation map. We propose  $\mathcal{L}_{P} = \mathcal{L}_{UPD} +  \mathcal{L}_{IPD}$.
The components of PLFD are as follows.

\textbf{Unified Pixel-level Distillation (UPD).} 
We propose distilling the precise predictive information from the teacher's unified pixel-wise predictive maps of middle layers, $p^t_{mid}$, to
the student’s early interpolated representation, $p^s_{early}$. 
\begin{align}
\mathcal{L}_{UPD} = \text{KL}(\mathcal{A}(p^{s}_{early}) || p^t_{mid})
\end{align}

\textbf{Individual Pixel-level Distillation.} Here, the teacher's terminal predictive map, denoted as $p^t_{late}$, distills precise information to the intermediate predicted maps of the students individually, represented as $p^{s}_{\text{mid}_j}$. This allows the student to capture detailed knowledge about the exact pixel locations and their corresponding class assignments within the image from much earlier layers. To achieve this, a KL-divergence loss is employed between these maps:
\begin{align}
\mathcal{L}_{IPD} &= \frac{1}{N} \sum^{N}_{j} \text{KL}(\mathcal{A}(p^{s}_{\text{mid}_j}) || p^t_{late})
\end{align}
Here, $N$ is the number of middle-layer blocks in the student. 

\subsection{Hierarchical Layer-selective Feedback Distillation (HLFD)}

Finally, distilling both feature-level and pixel-level representations allows the student to learn fine-to-coarse hierarchical details at both the feature and pixel levels. The multi-task loss function can be defined as:
\begin{align}
    \mathcal{L}_{H} =  \mathcal{L}_{Seg}  + \beta * \mathcal{L}_{F} + \lambda * \mathcal{L}_{P}
\end{align}

Where $\mathcal{L}_{seg}$ is the focal dice loss used for training the student network in a supervised fashion. In the inference phase, post-sufficient training, both the teacher network components and distillation modules are discarded.


\section{Experimental Platform}
\textbf{Datasets:} We evaluated our techniques on kidney tumor segmentation (KiTS)~\cite{heller2019kits19} and liver tumor segmentation (LiTS) ~\cite{bilic2023liver} datasets. KiTS comprises 210 abdominal CT scans, where a 168:42 split is used for testing and training. Similarly,  the LiTS dataset consists of 201 CT scans and uses the split of  131:70.

\textbf{Baselines:} We compare our method with the following SOTA methods: i) Structured Knowledge Distillation (SKD) \cite{liu2019structured}: Involves pair-wise distillation to capture similarity at feature and pixel level. ii) Intermediate Feature Distiller (IFD) \cite{ijcai2023p108}: Distills the teacher's terminal representation into concatenated branches of the student model. iii) Deep Knowledge Distillation (DKD) \cite{qin2021efficient}: Similar to \cite{qin2021efficient} but without the Relational Knowledge distillation(RKD) module. 
iv) Hierarchical Individual Feedback Knowledge Distillation (HIFD) \cite{luo2023knowledge}: It distills the teacher's terminal representation to individual layers of the student. We extended this method for segmentation by incorporating pixel-level feedback distillation loss functions.
We maintained identical implementation settings across all techniques.

\textbf{Implementation Details:} We employed UNet++~\cite{zhou2018unet++} architecture (36.1M parameters) as the teacher network and ResNet18~\cite{he2016deep} (11.6M parameters) as the student network. Our segmentation networks and distillation processes, inspired by~\cite{qin2021efficient}, were trained using Adam optimizer with beta1 (0.9) and beta2 (0.999). The learning rate began at 0.001, utilizing CosineAnnealing for rate scheduling, reaching a minimum of 0.000001. Data augmentation techniques such as random rotation and flipping were applied, while experiments revealed that Gaussian noise augmentation is unsuitable for medical images. Most networks processed authentic $512 \times 512$ CT images, requiring windowing of HU values with radiological standards (e.g., -40 to 160 for the liver and -200 to 300 for the kidney). We use the PyTorch framework. We train all the networks till convergence with up to 120 epochs. We report the result as $mean \pm std$ after three runs. For the Dice score (DSC), higher is better. For Relative Volume Difference (RVD), smaller absolute values are desired, indicating a closer match between the predicted and ground truth volumes. When comparing RVD values, a smaller absolute value (closer to zero) is better, regardless of whether the RVD is positive or negative. These metrics provide complementary insights about the performance.



\section{Results}
\textbf{Quantitative Results:}  
As shown in Table~\ref{table:1}, our method, HLFD, consistently outshines both the supervised student and the baseline models. Notably, on the KiTS dataset, we observe a substantial enhancement in DSC over the student (without KD). Further, both IFD and HIFD exhibit competitive or superior outcomes than other baselines. These results underscore the necessity of integrating KD and also emphasize the critical importance of architecting layers that adeptly distill the `what' and `where' dimensions of knowledge from the teacher model. On the RVD metric also, HLFD outperforms baselines, including the student (without KD), by a significant margin. This insight into volume differences holds valuable implications, especially in tasks like tumor segmentation where volume accuracy is of paramount importance.

\begin{table}[htbp]
\footnotesize
\centering
\caption{Quantitative Results ($\beta = 0.9$ and $\lambda = 0.1$) }
\label{table:1}
\begin{tabular}{c|cc|cc}
\toprule
\multirow{ 2}{*}{\textbf{Method}} & \multicolumn{2}{c} {\textbf{KiTS}} & \multicolumn{2}{c}{\textbf{LiTS}} \\
\cmidrule(lr){2-3} \cmidrule(lr){4-5}
&  \textbf{DSC} & \textbf{RVD} & \textbf{DSC} & \textbf{RVD} \\
\midrule
Teacher &  64.50 $\pm$ 1.45	& -0.203 & 57.84 $\pm$ 1.55 & 1.434\\
Student(w/o KD) &   41.30 $\pm$ 2.30	& -0.421 & 41.19 $\pm$ 1.65 & 0.701\\
\midrule
SKD \cite{liu2019structured} &  38.71 $\pm$  2.53 & -0.411 & 42.09 $\pm$ 2.01 & \textbf{0.018}\\
IFD \cite{ijcai2023p108}&  46.79 $\pm$  1.25	& -0.275 & 45.36 $\pm$ 1.20 & 0.122\\
DKD \cite{qin2021efficient}&  40.21 $\pm$  2.35	& -0.496 & 43.72 $\pm$ 0.87 & 0.234\\
HIFD \cite{luo2023knowledge}&  42.50 $\pm$  1.25	& -0.434 & 44.20 $\pm$ 1.22 & 0.187\\
\midrule
HLFD (ours) &  \textbf{52.18} $\pm$  2.55 & \textbf{-0.176} & \textbf{48.75} $\pm$ 2.23 & 0.173\\
\bottomrule
\end{tabular}
\end{table}

On the LiTS dataset, HLFD (our method) consistently outperforms the baselines on the DSC metric, whereas the SKD method is the best on the RVD metric. 

\textbf{Qualitative Results:} 
\begin{figure}[htbp]\centering
\def\svgwidth{\columnwidth}
\includegraphics[width=1.0\columnwidth]{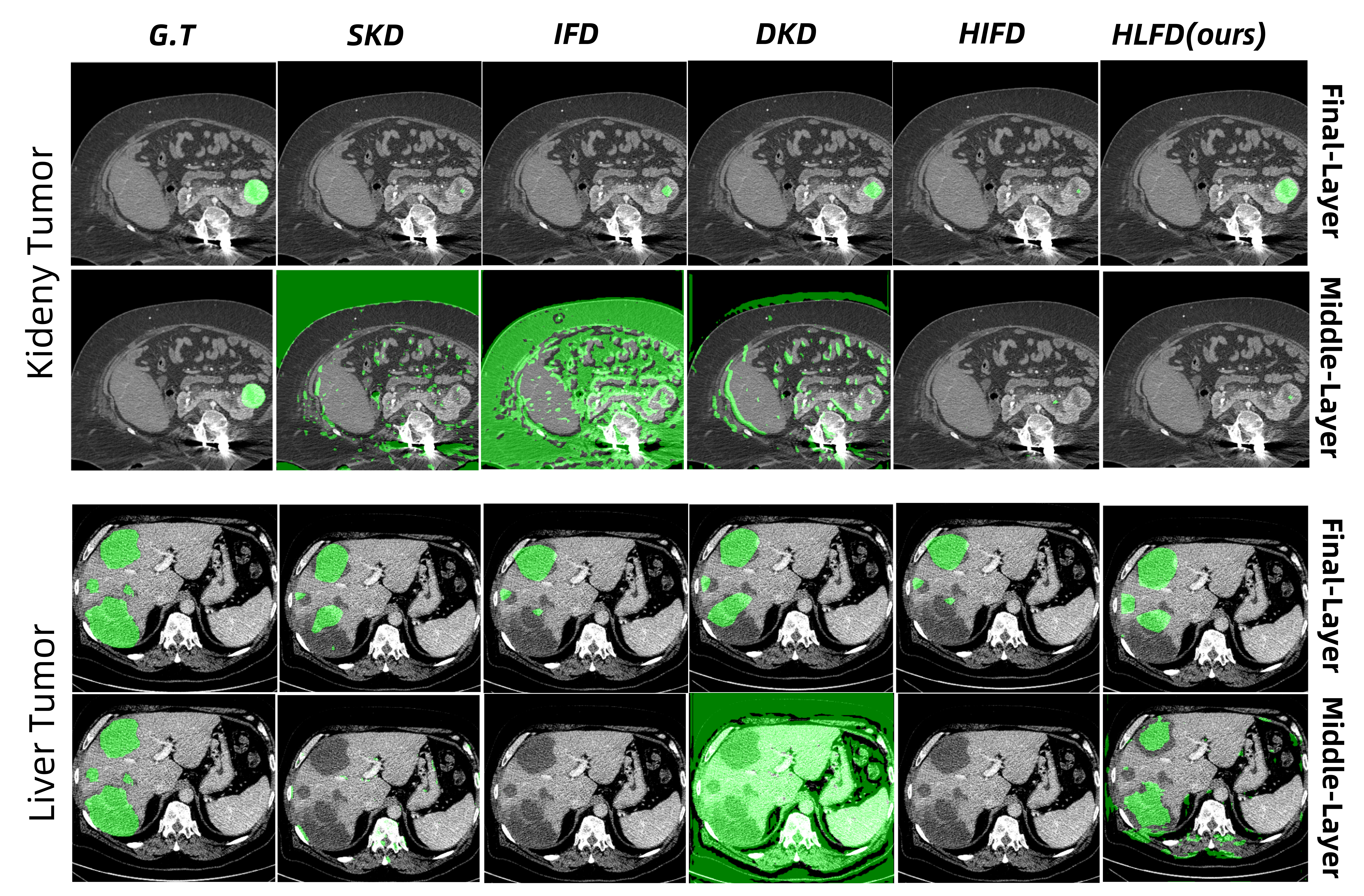}
\caption{The green color highlights the regions of interest (ROI) representing tumors. The segmentation maps are presented for both KiTS (first two rows) and LiTS (last two rows) datasets, with 'G.T.' denoting the ground truth.}
\label{pred:kits_lits}
\end{figure} 
The visualizations presented in Fig.~\ref{pred:kits_lits} showcase the superior performance of our proposed HLFD method. HLFD accurately segments the Region of Interest (ROI) while effectively suppressing irrelevant information, even at intermediate layers. The previous techniques fail to recognize the segmentation ROI at intermediate layers, especially for liver segmentation. This underscores the importance of meticulously designing layers to enhance the segmentation task's representation quality. 

\begin{figure}[htbp]
\centering
\includegraphics[width=1.0\columnwidth]{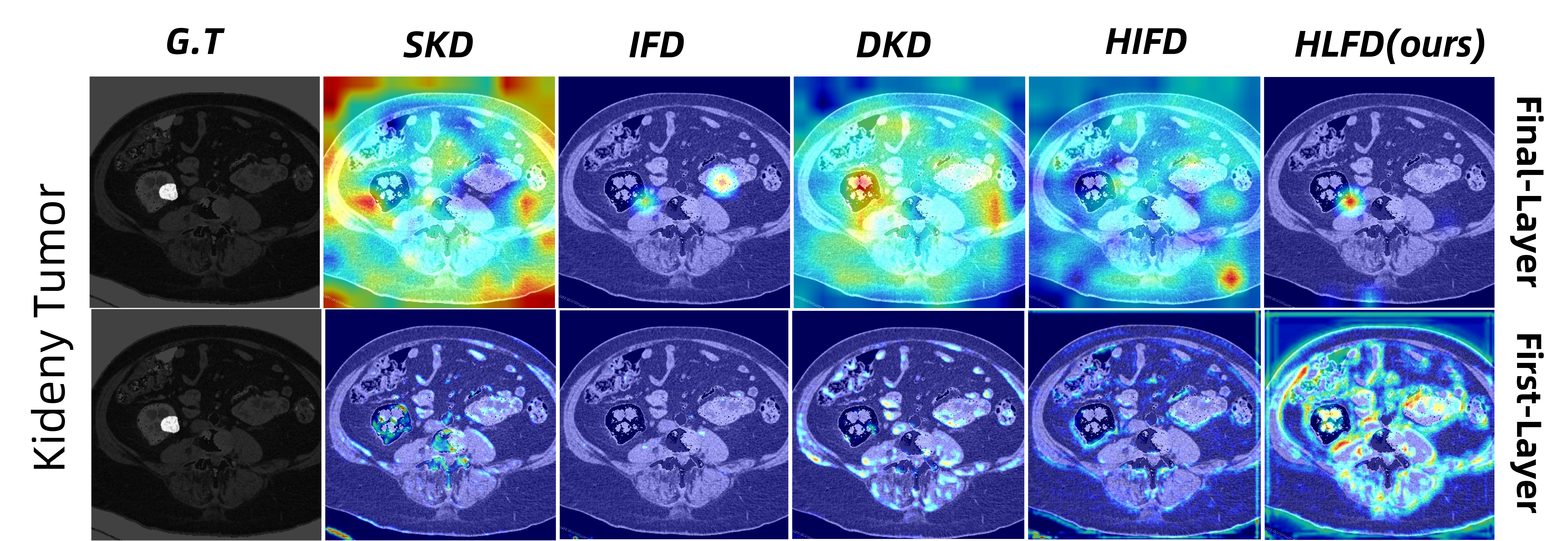}
\caption{Gradient-activated class maps for KiTS19, featuring CAM results for both the final and first layers. Remarkably, HLFD focuses on the tumor region, maintaining effectiveness in suppressing irrelevant information as the process advances to the final layer.}
\label{fig:cams}
\end{figure}

The GradCAM maps presented in Fig.~\ref{fig:cams} showcase distinct patterns among methods. SKD, which does not leverage intermediate layers, exhibits a flow of irrelevant information, hindering focus on tumor-specific details. While DKD shows some restriction of information, IFD and HIFD manage to suppress irrelevant details. However, they face challenges in focusing on tumor-specific information. In contrast, our method distinctly focuses on tumor-specific information without capturing irrelevant details.

\begin{table}[!ht]
\footnotesize
\centering
\caption{Impact of $\beta$ and $\lambda$ on DSC} \label{tab:sensitivity analysis}
\begin{tabular}{cc|c|c}
\toprule
${\beta}$ & ${\lambda}$ & \textbf{KiTS}  &\textbf{LiTS} \\
\midrule
0.9 & 0.1 & 52.18 $\pm$ 2.55 &  48.75 $\pm$ 2.23  \\
1.8 & 0.1 & 51.58 $\pm$ 1.25 &  48.18 $\pm$ 1.63  \\
0.9 & 0.2 & 51.95 $\pm$ 2.35 &  48.48 $\pm$ 2.05  \\
\bottomrule
\end{tabular}
\end{table}

\textbf{Sensitivity Analysis:} 
From Table~\ref{tab:sensitivity analysis}, doubling the $\beta$ value (from 0.9 to 1.8) while maintaining $\lambda$ constant led to a slight deterioration in performance. Conversely, increasing the value of $\lambda$ (from 0.1 to 0.2) while keeping $\beta$ constant showed a similar trend but with slightly improved performance compared to the previous case. This suggests that $\mathcal{L}_{F}$ learns a rich representation of data, while $\mathcal{L}_{P}$ learns the essential structure for the segmentation task. Therefore, maintaining $\beta$ greater than $\lambda$ is crucial for optimal results. The best performance was achieved with $\beta = 0.9$ and $\lambda = 0.1$.

\section{Conclusion}
We introduce a novel Knowledge Distillation (KD) framework for enhancing liver and kidney tumor segmentation, redefining knowledge selection to distill and the distillation source, and transitioning from teacher encoder layers to the student. Quantitatively, HLFD has demonstrated remarkable superiority over existing KD techniques and baseline models. Our method substantially improves DSC, particularly in kidney tumor segmentation, where HLFD surpasses the student model (without KD) by over 10\%. Qualitatively, HLFD exhibits exceptional capabilities in suppressing irrelevant information while maintaining a sharp focus on tumor-specific details. The ability of HLFD to accurately segment the region of interest (ROI) at both intermediate and final layers showcases its effectiveness in enhancing the quality of segmentation representations for kidney and liver tumor segmentation.

\section{Compliance with ethical standards}
\label{sec:ethics}
This research study was conducted retrospectively using human subject data made available in open access~\cite{emara2019liteseg,heller2019kits19}. Ethical approval was not required as confirmed by the license attached with the open-access data.

\section{Conflicts of Interest}
The authors have no relevant financial or non-financial interests to disclose.


\bibliographystyle{IEEEbib}
\bibliography{main}

\end{document}